\documentclass[10pt,twocolumn,letterpaper]{article}

\usepackage[accsupp]{axessibility}
\usepackage{wacv}              

\usepackage{graphicx}
\usepackage{amsmath}
\usepackage{amssymb}
\usepackage{booktabs}

\usepackage{tabularx}
\usepackage{placeins}
\usepackage{multirow}
\usepackage{float}

\usepackage[pagebackref,breaklinks,colorlinks]{hyperref}
\usepackage[numbers, sort]{natbib}

\usepackage[capitalize]{cleveref}
\crefname{section}{Sec.}{Secs.}
\Crefname{section}{Section}{Sections}
\Crefname{table}{Table}{Tables}
\crefname{table}{Tab.}{Tabs.}

\def\wacvPaperID{36} 
\def\confName{WACV}
\def\confYear{2024}

\begin{document}

\title{Filter-Pruning of Lightweight Face Detectors Using a Geometric Median Criterion \thanks{This work was supported by the EU Horizon 2020 programme under grant agreement H2020-951911 AI4Media. Code and trained pruned face detection models are available at: \url{https://github.com/IDT-ITI/Lightweight-Face-Detector-Pruning}} }

\author{Konstantinos Gkrispanis\\
CERTH-ITI\\
Thessaloniki, Greece, 57001\\
{\tt\small gkrispanis@iti.gr}
\and
Nikolaos Gkalelis\thanks{Work done while at CERTH-ITI.}\\
CERTH-ITI\\
Thessaloniki, Greece, 57001\\
{\tt\small gkalelis@iti.gr}
\and
Vasileios Mezaris\\
CERTH-ITI\\
Thessaloniki, Greece, 57001\\
{\tt\small bmezaris@iti.gr}
}
\maketitle

\begin{abstract}
Face detectors are becoming a crucial component of many applications, including surveillance, that often have to run on edge devices with limited processing power and memory. Therefore, there's a pressing demand for compact face detection models that can function efficiently across resource-constrained devices. Over recent years, network pruning techniques have attracted a lot of attention from researchers. These methods haven't been well examined in the context of face detectors, despite their expanding popularity.
In this paper, we implement filter pruning on two already small and compact face detectors, named EXTD (Extremely Tiny Face Detector) and EResFD (Efficient ResNet Face Detector). The main pruning algorithm that we utilize is Filter Pruning via Geometric Median (FPGM), combined with the Soft Filter Pruning (SFP) iterative procedure. We also apply L1 Norm pruning, as a baseline to compare with the proposed approach. The experimental evaluation on the WIDER FACE dataset indicates that the proposed approach has the potential to further reduce the model size of already lightweight face detectors, with limited accuracy loss, or even with small accuracy gain for low pruning rates.
\end{abstract}

\section{Introduction}
\label{sec:intro}

Face detection technology is the backbone of numerous advanced applications, including but not limited to surveillance \cite{Kumar2019}. It has undergone significant evolution in the past decade. Especially with the rise of edge computing, where computations are performed on local devices with minimum computational power, efficient and compact face detectors have become necessary. While there is a need for these models to be lightweight, so they can run on edge devices, they shouldn't compromise on accuracy.

\begin{figure}[!htb]
\begin{center}
\begin{tabular}{cc}
\includegraphics[width=0.44\columnwidth]{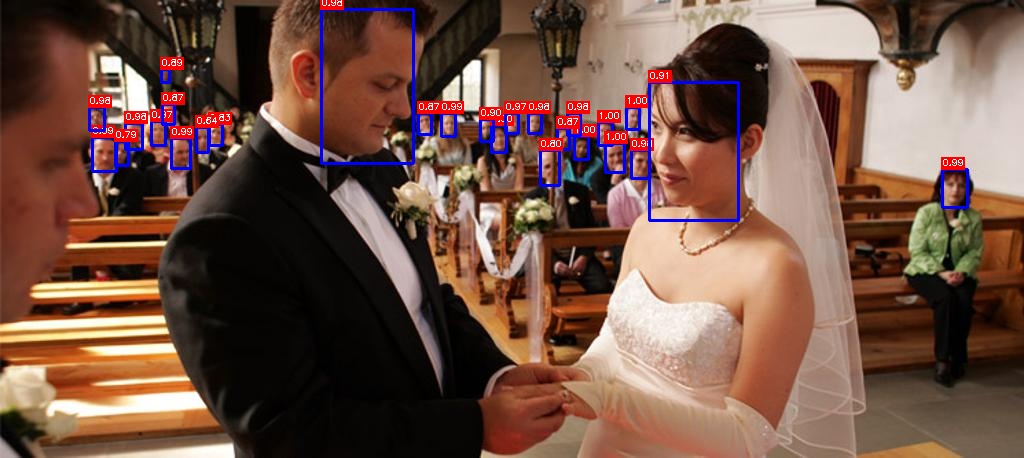} &
\includegraphics[width=0.44\columnwidth]{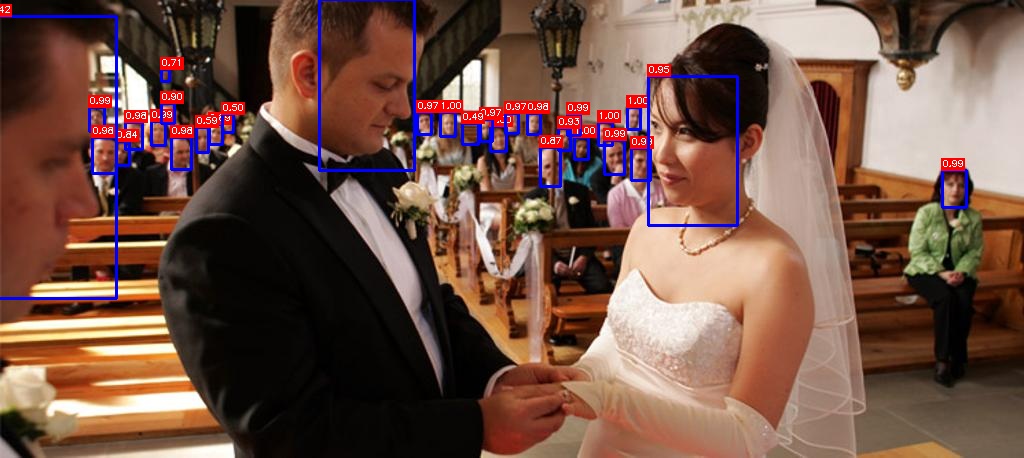} \\
{\small Original model} & {\small Pruned model} 
\end{tabular}
\end{center}
\caption{The proposed approach is used to prune EResFD (an already very lightweight face detector) with 10\% pruning rate. In this example, we see that the pruned model not only is more compact, but also detects faces, which were not detected with the original model.
This may be attributed to the regularization effects of our approach for small pruning rates.  
}
\label{fig:PruningExample}
\end{figure}

One promising avenue to achieve this balance is through network pruning \cite{248452}. Network pruning is a technique aimed at reducing the size of deep learning models without a significant drop in their performance. Over the years, various pruning techniques have been proposed and have achieved considerable success in tasks like image classification. Yet, their application and potential benefits in the domain of face detection remain largely uncharted.
Specifically, to the best of our knowledge, only the method in \cite{10.1145/3436369.3437415} utilizes a pruning approach to a face detection network.
However, in the above work a criterion is used to prune the ``least important'' filters in the layer, which is not always optimal \cite{he2019filter}.

Considering the above, this paper aims to examine the application of network pruning to face detectors, especially to the most lightweight architectures of them.
Specifically, we adapt the Filter Pruning via Geometric Median (FPGM) \cite{he2019filter}  pruning algorithm and the Soft Filter Pruning (SFP) iterative procedure \cite{he2018soft} to prune two already compact and small, in terms of parameters, face detectors, namely, EXTD (Extremely Tiny Face Detector) \cite{yoo2019extd} and EResFD (Efficient ResNet Face Detector) \cite{jeong2022eresfd}.
FPGM identifies and prunes the filters with the ``most redundancy'', a principle that has shown to provide improved performance over other pruning algorithms in the literature.
Additionally, as baseline we compare with the widely used L1 Norm pruning criterion \cite{Kumar2021}, as representative of the ``less important'' pruning principle.
The experimental results on the WIDER FACE dataset \cite{yang2016wider} shows that the proposed approach has the potential to provide even more compact face detectors with competitive detection performance, especially when small pruning rates are used (e.g. see Figure \ref{fig:PruningExample}).

In overall, we aim to provide a comparative analysis of the above algorithms and determine which of them, if any, offers a notable advantage in the face detection context.
Through our research, we aspire to lay the foundations for more efficient and compact face detectors suitable for deployment on edge device, thereby broadening the horizons for real-time, resource constrained applications.
In summary, we make the following contributions:
\begin{itemize}
  \item We are the first to apply a redundancy-based pruning algorithm (FPGM) in order to prune the most redundant filters in lightweight face detection networks.  
  \item The proposed approach yields a family of even more lightweight face detection networks that achieve superior detection accuracy in comparison to the previous state-of-the-art models of similar size.
\end{itemize}

The rest of the paper is structured as follows:
The related work and proposed methodology are presented in Sections \ref{sec:related}
and \ref{sec:proposed}, respectively. Experimental results are discussed in Section \ref{sec:experiments} and conclusions are drawn in Section \ref{sec:conclusions}.

\section{Related Work}
\label{sec:related}

\subsection{Face Detection}

In recent times, Convolutional Neural Networks (CNN) and other deep learning architectures, such as Transformers, have achieved notable success in a variety of computer vision tasks, including image classification, object detection and semantic segmentation.

Face detection, a sub-task of object detection, similarly benefits from the effectiveness of CNNs \cite{Alzubaidi2021,Kumar2019}. Although Transformer-based architectures have shown state-of-the-art performance on object detection, the majority of state-of-the-art face detectors are extensions of CNN-based general-purpose object detectors \cite{minaee2021going}. PyramidBox \cite{tang2018pyramidbox} is based on Single-Shot Detector (SSD) \cite{liu2016ssd}. RetinaFace \cite{deng2019retinaface}, also based on SSD, is a robust single-stage face detector that performs pixel-wise face localization on various scales by leveraging both extra-supervised and self-supervised multi-task learning. TinaFace \cite{zhu2020tinaface} is treating face detection as a one-class generic object detection and is using Deformable Convolution, Intersection over Union (IoU) aware branches and a Distance IoU Loss to enhance the model's capability. YOLO5Face \cite{qi2022yolo5face} is based on a set of state-of-the-art object detection models, named YOLO (You Only Look Once) known for their real-time processing capabilities. S3FD \cite{zhang2017s3fd} is designed to efficiently detect faces across various scales, especially small ones, and achieves it by employing a scale-equitable framework (using anchors on a wide range of layers), a scale compensation anchor matching strategy and a max-out background. Dual Shot Face Detector (DSFD) \cite{li2019dsfd} addressed the challenges in face detection through three main contributions: Feature Enhance Module, Progressive Anchor Loss and Improved Anchor Matching. Most of the above-discussed models are based on relatively heavy ResNet-50 and VGG16 networks. Thus, they are not suitable for resource-constrained environments, such as edge devices.

\subsection{Lightweight Face Detectors}

Designing lightweight face detectors that can operate on edge devices, and other resource-constrained environments, is an active research area. For instance, RetinaFace \cite{deng2019retinaface} has provided a lightweight version, implemented  using MobileNet \cite{howard2017mobilenets} as a backbone. Multi-task convolutional neural network (MTCNN) \cite{8110322}, used Multi-task Cascaded Convolutional Networks and employed a three stage cascade structure to predict face and landmark locations. Faceboxes \cite{zhang2017faceboxes} is based on SSD \cite{liu2016ssd}; it combines Rapidly Digested Convolutional layers (RDCL) for swift input processing and Multiple Scale Convolutional Layers (MSCL) to handle faces of varying sizes. The authors in SCRFD \cite{guo2021sample}, utilizes Sample Redistribution to augment training samples and Computation Redistribution to reallocate computational resources across the model's backbone and improve the computational efficiency.
In \cite{he2019lffd}, an anchor-free one-stage face detection method optimized for edge devices is presented.
EXTD, proposed in \cite{yoo2019extd}, is able to detect faces at multiply scales by iteratively reusing a lightweight backbone network.
In \cite{jeong2022eresfd}, EResFD emphasizes the effectiveness of standard convolution for lightweight face detection.
That is, instead of using depthwise separable convolution (as in e.g. \cite{yoo2019extd}), it is demonstrated that a ResNet with significantly reduced channels in combination with a standard convolution can achieve similar results.
While many of the aforementioned detectors introduced novel techniques and achieved competitive performance on the WIDER FACE dataset \cite{yang2016wider}, the latter two models have demonstrated superior performance in terms of a good trade-off between accuracy and model size.
Consequently, in this study, we focus on EXTD  and EResFD since, to the best of our knowledge, they offer the best accuracy-to-parameters-used trade-off. Our aim is to derive even smaller and more compact models without a significant drop in performance.

\subsection{Network Pruning}

Network pruning approaches can be roughly categorized to structured and non-structured.  The latter, remove single weights, resulting in irregular weight sparsities, and thus require the use of specialized software and hardware to allow the efficient deployment of the pruned models.
On the other hand, structured pruning methods remove entire model components, such as filters,
yielding models that can be easily deployed.
For this reason, structured pruning is receiving greater attention in the community \cite{li2016pruning, Luo_2017_ICCV}. 

Due to the advantages described above, in this work we choose to perform structured pruning of very lightweight face detectors, and more specifically filter pruning.
While filter pruning has been intensively investigated in several image classification tasks, the pruning of face detectors is a relatively unexplored topic.
PruneFaceDet \cite{10.1145/3436369.3437415} is one of the few approaches in this domain.
It employs a L1 reqularization penalty imposed on the scaling factors of the Batch Normalization (BN) layers to perform structured pruning on the EagleEye face detector \cite{s19092158}.
Due to this fact, this method can be only used on networks of specific structure, i.e., a one-to-one association of convolutional and BN layers is required.
Additionally, pushing the values of the BN scaling factors towards zero, this approach is based on the ``least importance'' pruning principle.
In contrary, here we use the FPGM pruning algorithm that formulates pruning from a redundancy reduction perspective, which has shown superior performance in several image classification tasks \cite{he2019filter,9327924}.

\section{Proposed Methodology}
\label{sec:proposed}

Consider an individual convolutional layer in a face detector with weight parameters,
\begin{equation}
    F = [ F_1, \dots, F_n], \label{e:convlayer}
\end{equation}
where, $F_j \in \mathbb{R}^{k \times k \times c}$ is the $j$th filter with spatial size $k \times k$ and depth $c$, and $n$ is the total number of filters in the layer.
Based on the above formulation, the goal of the proposed approach is: given a filter pruning rate $\theta$ (common for all filters) prune the $n \theta$ filters in each layer of the face detector.

\subsection{Backbone Networks}
\label{ssec:backbone}

The first model that we choose to prune is the already compact EXTD \cite{yoo2019extd}, a state-of-the-art multi-scale face detector with an exceptionally small number of parameters. Unlike traditional multi-scale face detection models that extract feature maps of varying scales from a single backbone network, EXTD generates these feature maps by iteratively reusing a shared lightweight and shallow backbone network. This iterative sharing significantly reduces the model's parameters and provides abstract image semantics from higher network layers to the lower-level feature map. The key innovation is the ability to share the network in generating each feature map, which not only reduces the number of parameters but also enables the model to use more layers for detecting small faces. The architecture can be applied to both SSD and FPN (Feature Pyramid Network) based detection structures. Through experiments, it was demonstrated in \cite{yoo2019extd} that this model can handle faces of various scales and conditions. 

The second model we choose to prune is EResFD \cite{jeong2022eresfd}. Here, it is shown that the combination of reduced channels with standard convolution can achieve similar results. EResFD consists of a modified ResNet backbone and feature enhancement modules: Separated Feature Pyramid Network and Cascade Context Prediction Module. To the best of our knowledge, this represents the face detector with the fewest parameters currently available, approximately 90,000 parameters. Thus, further pruning poses a significant challenge.

\subsection{Pruning Algorithms}
\label{ssec:criterion}

Redundancy-based pruning algorithms, i.e. algorithms that identify and discard the filters in a layer with the most similar characteristics, have shown superior performance in comparison to other criteria in the literature \cite{he2019filter,9327924}.
To this end, we resort to the FPGM algorithm, which has been successfully used to prune different types of backbone networks and for different applications \cite{he2019filter,9327924}. Additionally, for comparison purposes, the well-known L1 Norm algorithm is utilized  as our baseline. Both algorithms are briefly described in the following:

\begin{itemize}
     \item \textbf{L1 Norm}: The L1 Norm pruning algorithm, as proposed in \cite{li2016pruning} and applied on various problems, e.g. \cite{Kumar2021}, focuses on evaluating the significance of groups of weights (such as filters) in convolutional layers based on their L1 norm. It is based on the traditional "smaller-norm-less-important" criterion. 
     Given filter $F_j$ (\ref{e:convlayer}) its L1 norm is computed as:
    \begin{equation}
    \text{norm}(F_j, 1) = \sum_{i=1}^{ck^2} |f_{i,j}|,
    \end{equation}
    where, $f_{i,j}$ is the $i$th element of $F_j$.
    Filters with smaller L1 norm values, which indicate lower overall importance, are pruned. This methodology is inspired by the intuition that smaller-norm weight filters contribute less to the model's final prediction.

    \item \textbf{FPGM}: This algorithm has been originally proposed in \cite{he2019filter}.
    It is based on the Geometric Median (GM),
    the classic robust estimator of centrality for data in Euclidean spaces, to prune redundant filters in a convolutional layer.
    Contrary to the L1 Norm algorithm, it aims to identify filters that carry redundant information.
    As defined in \cite{he2019filter}, the GM of a set of filters, denoted as $x^{\text{GM}}$, is mathematically expressed as:
    \begin{equation}
    x^{\text{GM}} = \arg\min_{x \in \mathbb{R}^{ {k \times k \times c} }} \sum_{j' \in [1, {n}]} \|x - F_{j'}\|_2,
    \end{equation}
    where, $x$ represents a filter in a layer and is used as a placeholder to denote any filter from the set of filters in that layer.
    Subsequently, the filter in the layer closest to this geometric median is given by:
    \begin{equation}
    F_{j^*} = \arg\min_{F_{j'}} \|F_{j'} - x^{\text{GM}}\|_2, \text{ s.t. } j' \in [1, n].
    \end{equation}
    To mitigate the computational cost of finding the GM, the following algorithm that identifies the filter minimizing the aggregate distance to all other ones is used:
    \begin{equation}
    F_{x^*} = \arg\min_{x} g(x), \text{ s.t. } x \in \{ F_{1}, \ldots, F_{n} \},
    \end{equation}
    where the function $g(x)$ is defined as:
    \begin{equation}
    g(x) = \sum_{j'=1}^{n} \| x - F_{j'} \|_2.
    \end{equation}
    The filter \( F_{x^*} \) that minimizes \( g(x) \) can be then pruned with minimal impact on the network's redundancy.
    Experimental results in \cite{he2019filter} validate the efficacy of FPGM, showing significant performance improvements on CIFAR-10 and ILSVRC2012 datasets.
    While the FPGM algorithm has been validated in several classification tasks (e.g. see \cite{sym12091426}, \cite{9327924}), its utility in face detection is yet to be explored.
\end{itemize}

\subsection{Soft Filter Pruning}
\label{ssec:soft_pruning}

We combine the FPGM algorithm with the Soft Filter Pruning (SFP) procedure  \cite{he2018soft} to prune the face detector in an iterative manner.
Contrary to other methods in the literature that permanently prune filters, SFP allows pruned filters to be updated during subsequent model training.
One main advantage offered by this approach is that it retains a larger model capacity since updating previously pruned filters provides a broader optimization space compared to permanently setting filters to zero.
This larger optimization space allows the network to better learn from training data.

\section{Experiments}
\label{sec:experiments}

\subsection{Dataset and Metrics}
\label{ssec:dataset_metrics}

The dataset used for training and evaluation in this work is the WIDER FACE dataset \cite{yang2016wider}, a widely-used dataset for face detection research. It contains 32,203 images and embraces a wide variety of challenges, including large variations in scale, pose and occlusion. It is structured based on 60 event classes and the faces within it demonstrate significant variability in appearance, making it a challenging benchmark. Based on the the level of difficulty of the faces to be detected, the images are categorized into three subsets: Easy, Medium and Hard.

The performance evaluation is performed using the Mean Average Precision (mAP).
We should note that this metric is typically used in conjunction with the WIDER FACE dataset to evaluate the performance of the different models across all three subsets.

\subsection{Experimental Settings}
\label{ssec:experimental_settings}

Both EXTD \cite{yoo2019extd} and EResFD \cite{jeong2022eresfd}, were evaluated using the following pruning rates $\theta = \{0.1, 0.2, 0.3, 0.4, 0.5$\}.
The pruning rate refers to sparsity per pruned layer.
All convolutional layers of the models are chosen to be pruned, except of those that are part of the detection head.
The detection head is the last component of the model and is responsible for predicting the final bounding boxes and the classification of them - whether they contain a face or not; it is a crucial part of the network, hence we decided against pruning it.

The initial EXTD and EResFD models were created after we trained EXTD and EResFD from scratch with the setups that were described in their original papers.
Subsequently, the  iterative process of SFP  \cite{he2018soft} combined with the pruning algorithm (FPGM or L1 Norm) was conducted over 200 epochs to prune the different models.
Specifically, a step learning rate schedule was adopted, with an initial learning rate of 1e-3.
This was scaled down by a factor of 0.1 at epochs 50 and 100.
For optimizer selection in each experiment, we strictly followed the configurations reported in the original publications.
That is, the Stochastic Gradient Descent (SGD) with momentum 0.9, and the Adam optimizer, both with weight decay 5e-4 were used for the EXTD and EResFD, respectively.
Upon reaching epoch 200, the SFP was halted, and further fine-tuning was conducted without updating the pruned weights during backpropagation (i.e. the pruned weights remained at zero).
For this fine-tuning step, the learning rate was reset to 1e-3 for 5 epochs, followed by a decay to 1e-4 for the remaining 5 epochs of the process.

All the models and algorithms were implemented in PyTorch \cite{paszke2019pytorch}.
For the implementation of the FPGM and L1 Norm algorithms the NNI library \cite{nni2021} functions \verb|FPGMPruner| and \verb|L1NormPruner| were used, respectively.

\begin{figure*}[htbp]
    \centering
    \includegraphics[width=0.96\linewidth]{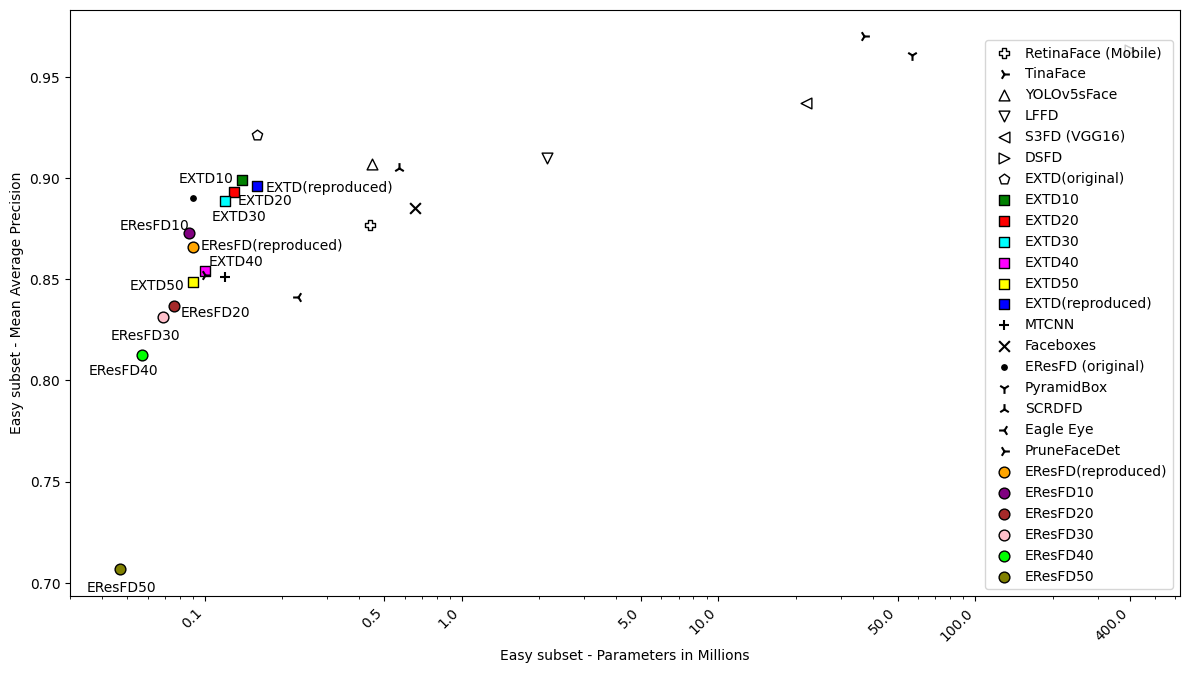}
    \caption{Model size and mAP across different face detectors on the Easy subset of WIDER FACE.}
    \label{fig:fig1}
\end{figure*}

\begin{figure*}[htbp]
    \centering
    \includegraphics[width=0.96\linewidth]{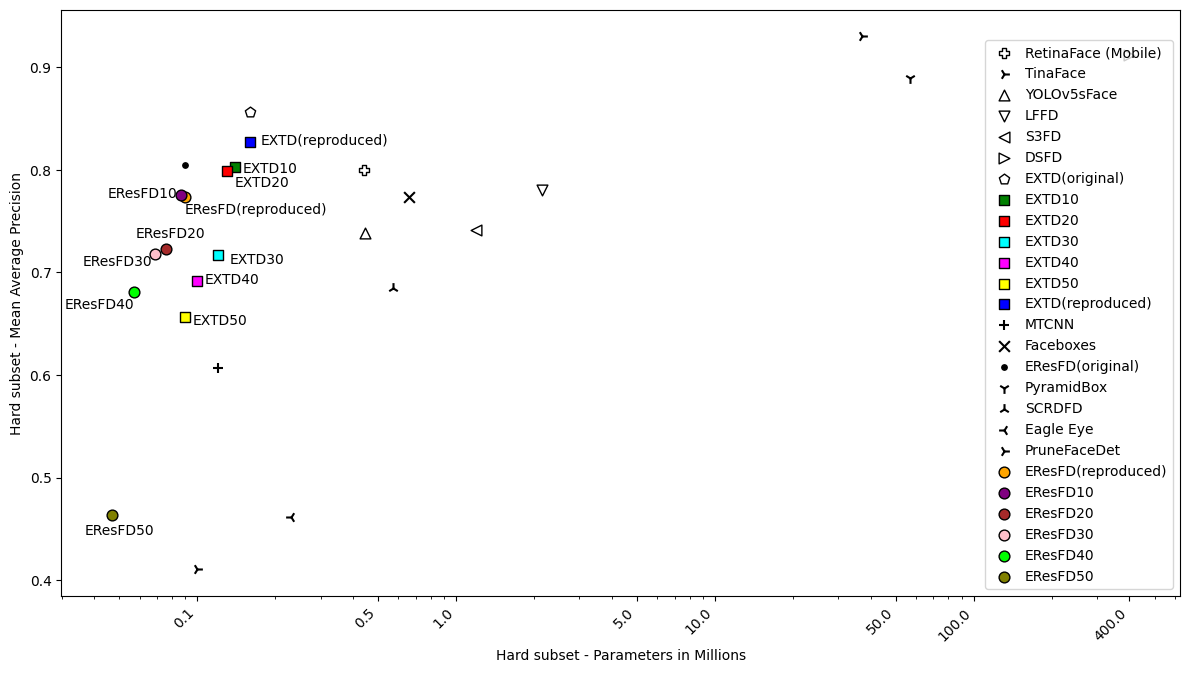}
    \caption{Model size and mAP across different face detectors on the Hard subset of WIDER FACE.}
    \label{fig:fig2}
\end{figure*}

\begin{table*}
\centering
\caption{Comparative results on EXTD between the proposed pruning approach (FPGM-based) and our baseline.}
\label{tab:sfp_on_extd}
\small 
\begin{tabular}{l|ccc|c|c}
\toprule
Method & Easy & Medium & Hard & \centering \# of Parameters & \centering\arraybackslash Real Sparsity \\
\midrule
EXTD(original, from \cite{yoo2019extd}) & 0.9210 & 0.9110 & 0.8560 & \multirow{2}{*}{162,352} & \multirow{2}{*}{0\%} \\
EXTD (reproduced) & 0.8961 & 0.8868 & 0.8268 & & \\
\midrule
FPGM 10\% & 0.8988 & 0.8828 & 0.8026 & \multirow{2}{*}{149,472} & \multirow{2}{*}{7.93\%} \\
L1 10\% & 0.8950 & 0.8766 & 0.7961 & & \\
\midrule
FPGM 20\% & 0.8931 & 0.8789 & 0.7992 & \multirow{2}{*}{136,296} & \multirow{2}{*}{16.05\%} \\
L1 20\% & 0.8921 & 0.8766 & 0.7923 & & \\
\midrule
FPGM 30\% & 0.8885 & 0.8588 & 0.7168 & \multirow{2}{*}{122,034} & \multirow{2}{*}{24.83\%} \\
L1 30\% & 0.8806 & 0.8522 & 0.6655 & & \\
\midrule
FPGM 40\% & 0.8539 & 0.8213 & 0.6915 & \multirow{2}{*}{108,858} & \multirow{2}{*}{32.95\%} \\
L1 40\% & 0.8427 & 0.8068 & 0.6544 & & \\
\midrule
FPGM 50\% & 0.8485 & 0.8118 & 0.6565 & \multirow{2}{*}{94,448} & \multirow{2}{*}{41.83\%} \\
L1 50\% & 0.8422 & 0.7971 & 0.6267 & & \\
\bottomrule
\end{tabular}
\end{table*}

\begin{table*}
\centering
\caption{Comparative results on EResFD between the proposed pruning approach (FPGM-based) and our baseline.}
\label{tab:sfp_on_eres}
\small 
\begin{tabular}{l|ccc|c|c}
\toprule
Method & Easy & Medium & Hard & \centering \# of Parameters & \centering\arraybackslash Real Sparsity \\
\midrule
EResFD(original, from \cite{jeong2022eresfd}) & 0.8902 & 0.8796 & 0.8041 & \multirow{2}{*}{92,208} & \multirow{2}{*}{0\%} \\
EResFD (reproduced) & 0.8660 & 0.8555 & 0.7731 & & \\
\midrule
FPGM 10\% & 0.8728 & 0.8582 & 0.7757 & \multirow{2}{*}{87,368} & \multirow{2}{*}{5.25\%} \\
L1 10\% & 0.8470 & 0.8345 & 0.7410 & & \\
\midrule
FPGM 20\% & 0.8369 & 0.8201 & 0.7230 & \multirow{2}{*}{76,677} & \multirow{2}{*}{16.84\%} \\
L1 20\% & 0.8263 & 0.8038 & 0.6723 & & \\
\midrule
FPGM 30\% & 0.8311 & 0.8160 & 0.7175 & \multirow{2}{*}{69,746} & \multirow{2}{*}{24.36\%} \\
L1 30\% & 0.8218 & 0.8001 & 0.6663 & & \\
\midrule
FPGM 40\% & 0.8124 & 0.7952 & 0.6807 & \multirow{2}{*}{57,055} & \multirow{2}{*}{35.95\%} \\
L1 40\% & 0.7603 & 0.7349 & 0.5800 & & \\
\midrule
FPGM 50\% & 0.7103 & 0.6830 & 0.5254 & \multirow{2}{*}{47,284} & \multirow{2}{*}{48.72\%} \\
L1 50\% & 0.6992 &  0.6704 & 0.4824 & & \\
\bottomrule
\end{tabular}
\end{table*}

\begin{figure*}[!htb]
\begin{center}
\begin{tabular}{ccc}
EXTD & EXTD10 & EXTD50 \\
\includegraphics[width=0.27\linewidth]{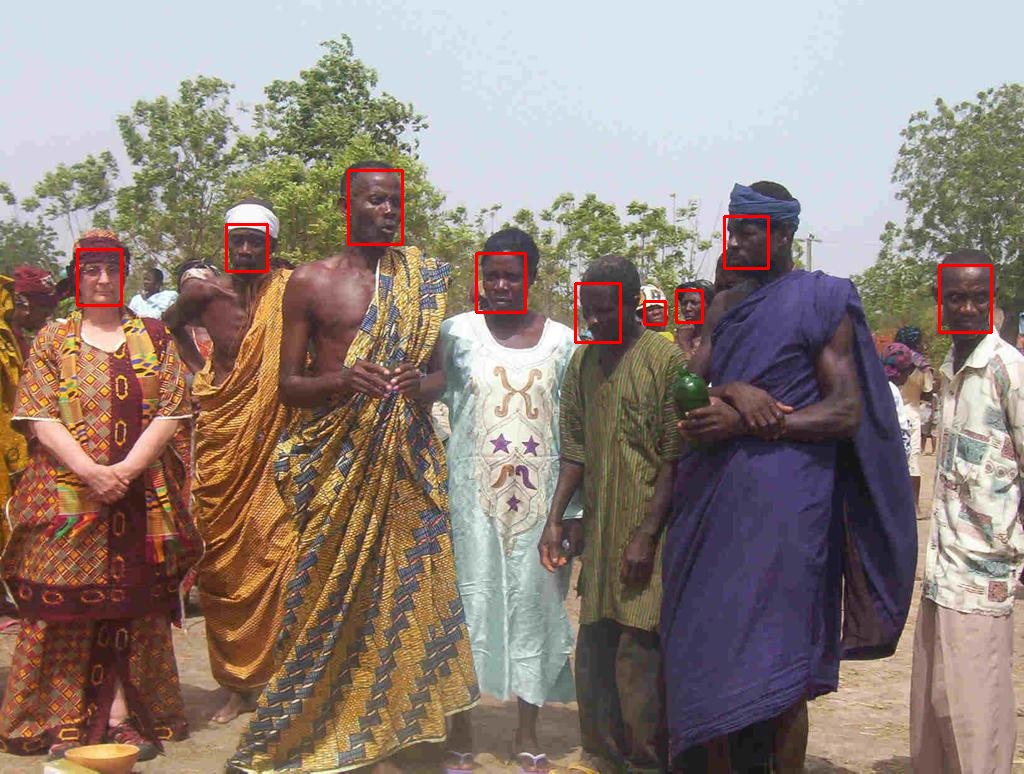} &
\includegraphics[width=0.27\linewidth]{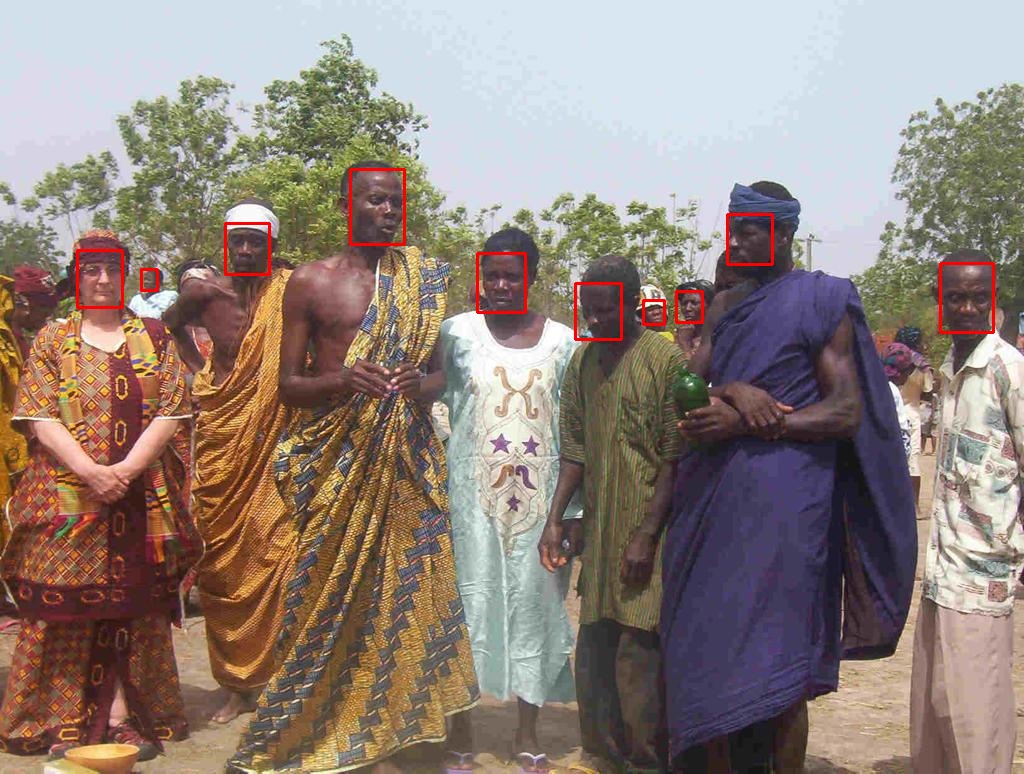} &
\includegraphics[width=0.27\linewidth]{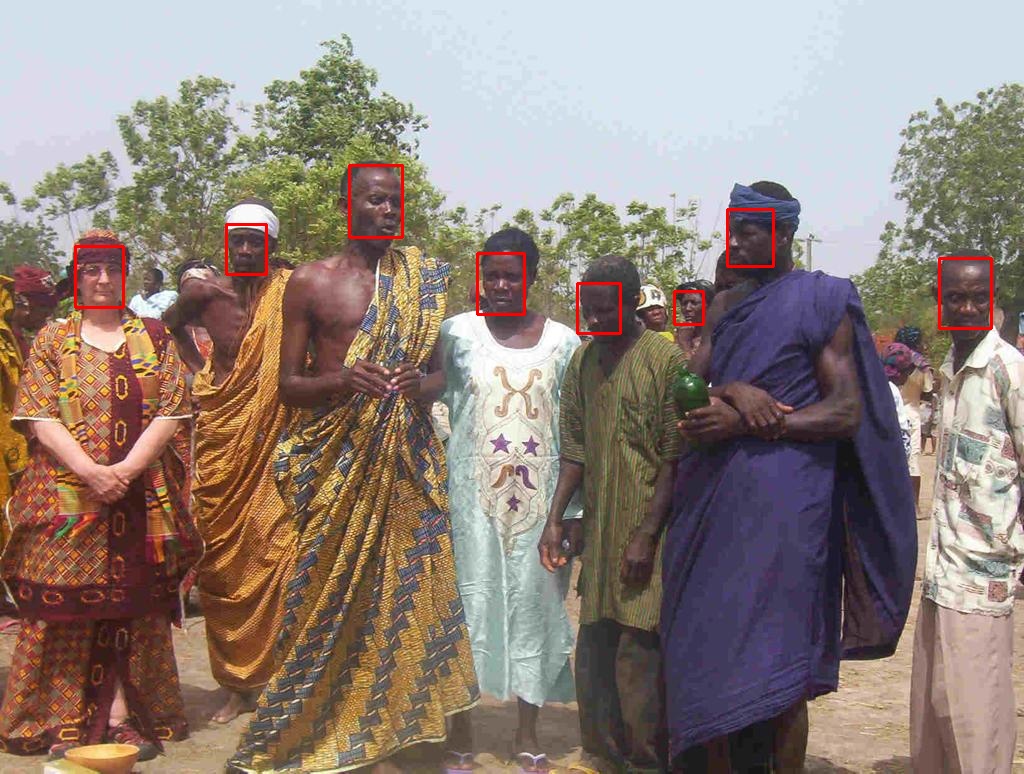} \\
\includegraphics[width=0.27\linewidth]{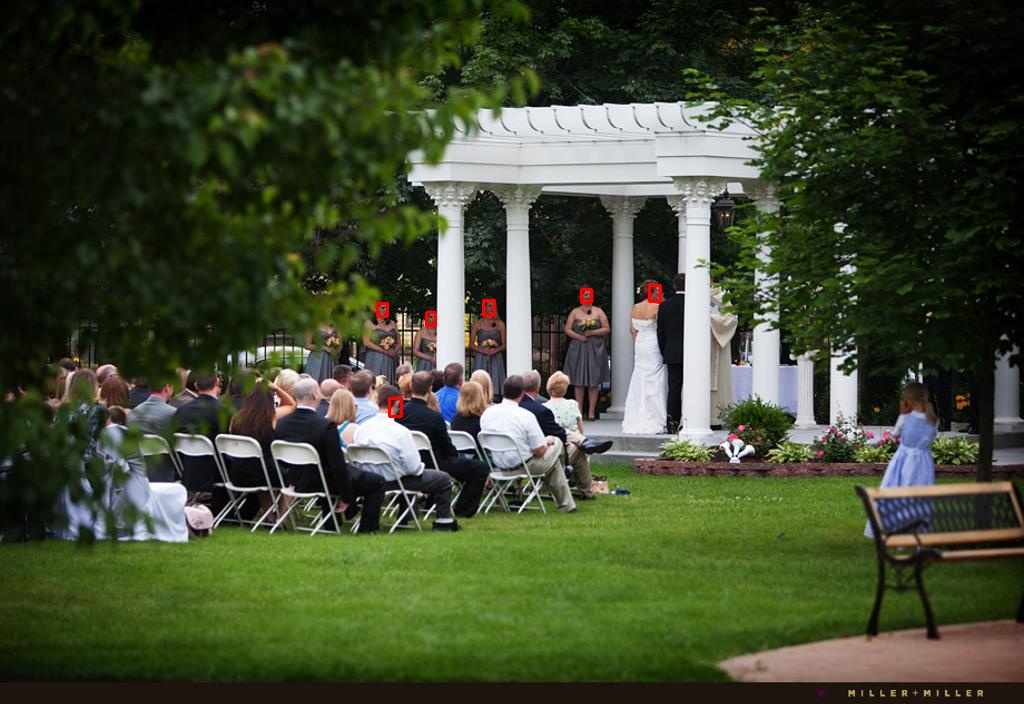} &
\includegraphics[width=0.27\linewidth]{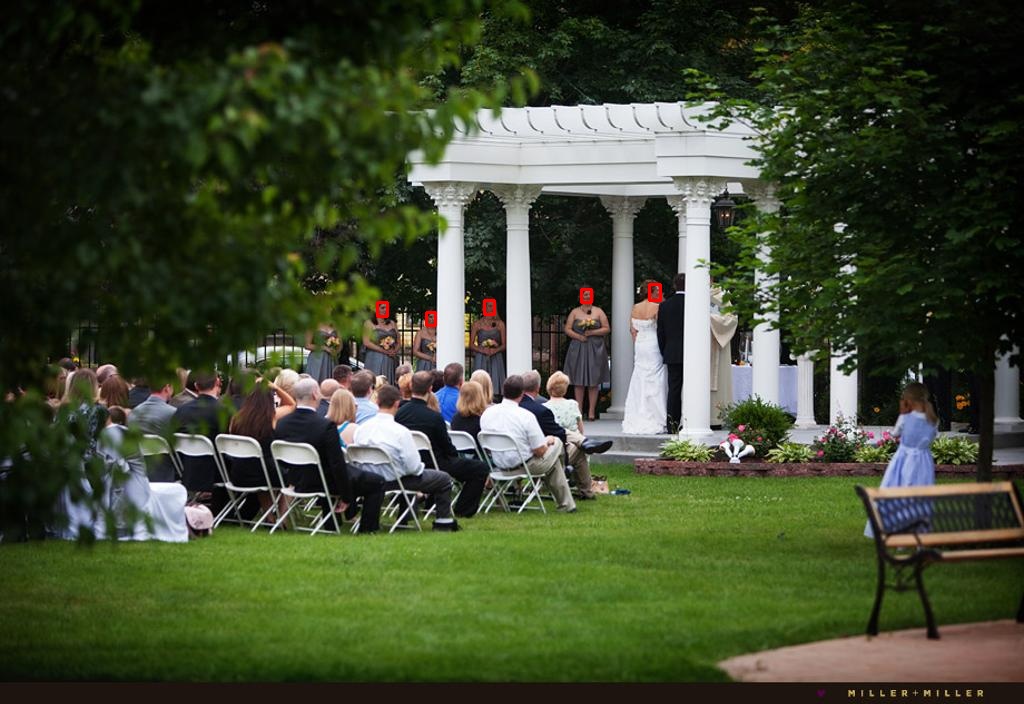} &
\includegraphics[width=0.27\linewidth]{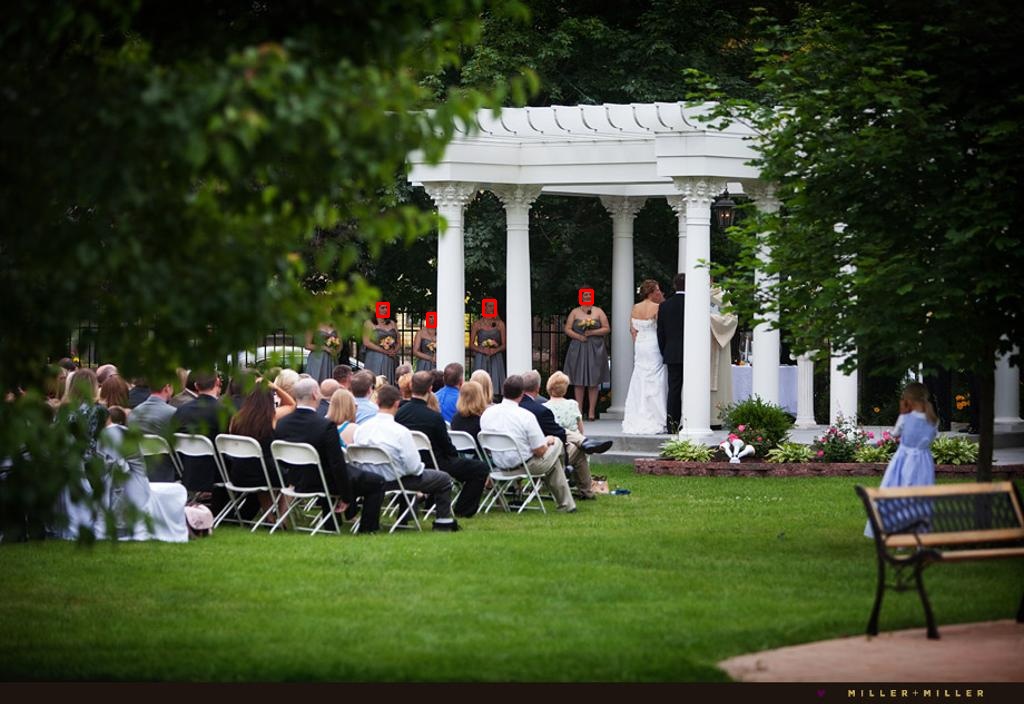} \\
\includegraphics[width=0.27\linewidth]{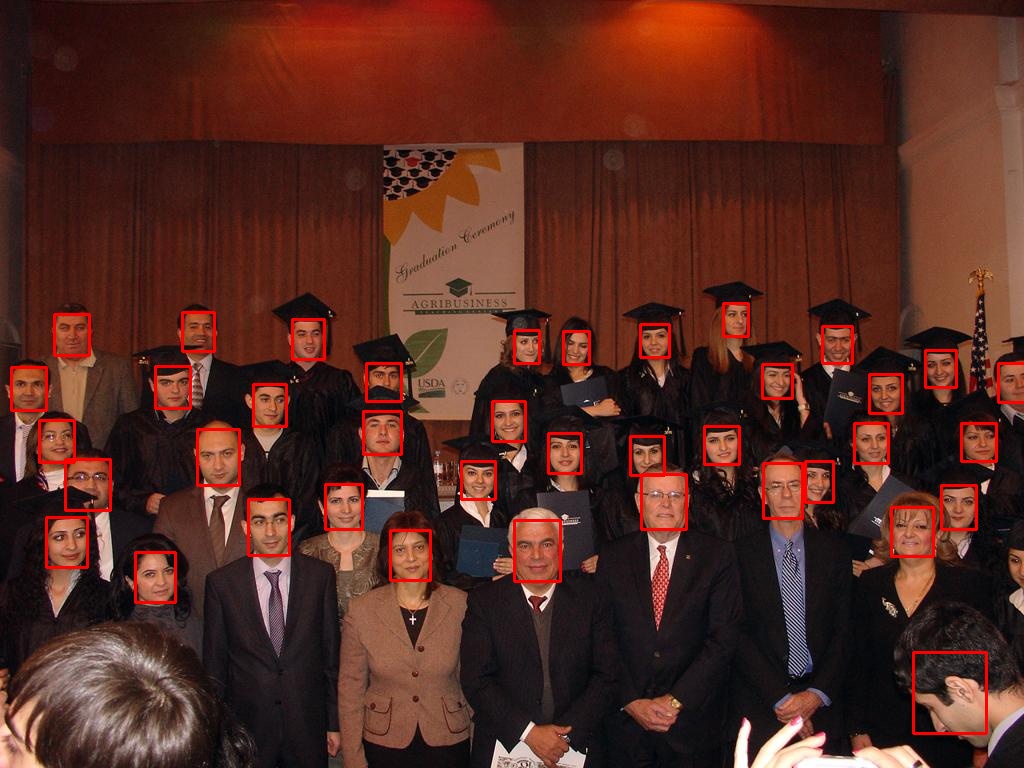} &
\includegraphics[width=0.27\linewidth]{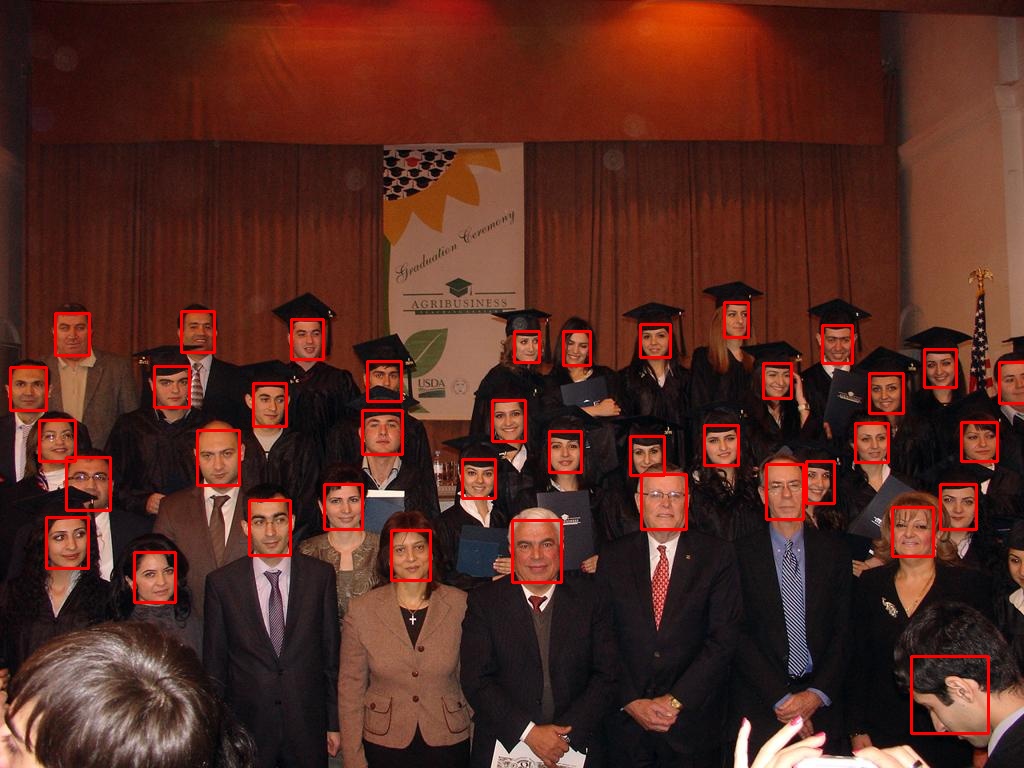} &
\includegraphics[width=0.27\linewidth]{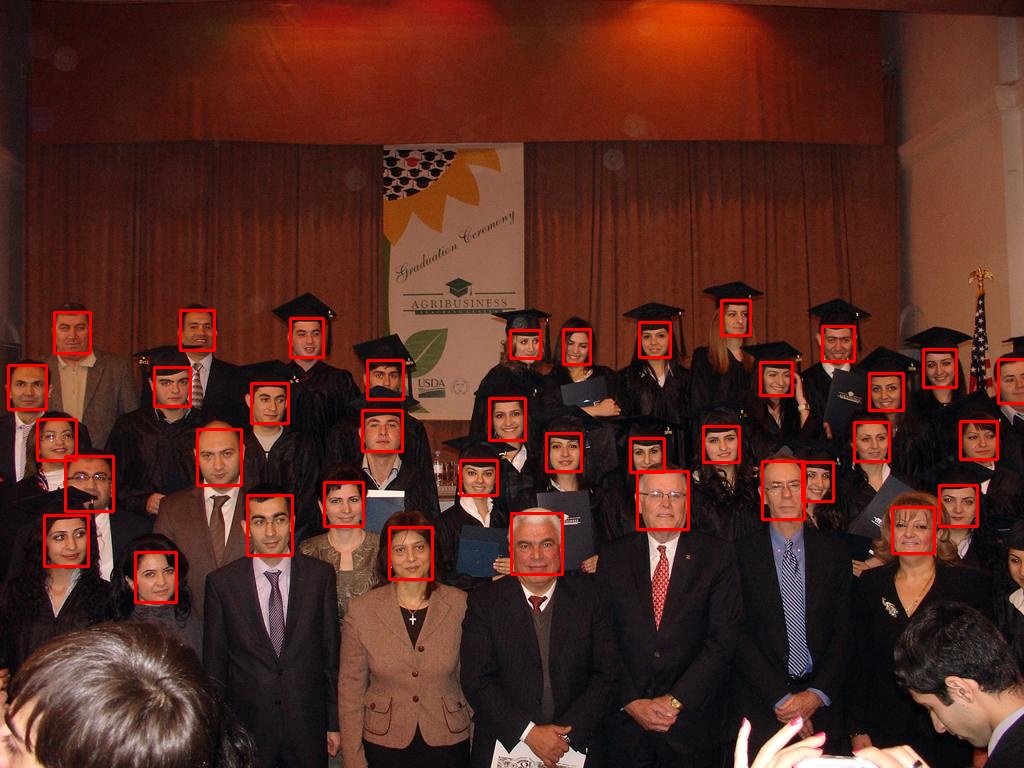} \\
\includegraphics[width=0.27\linewidth]{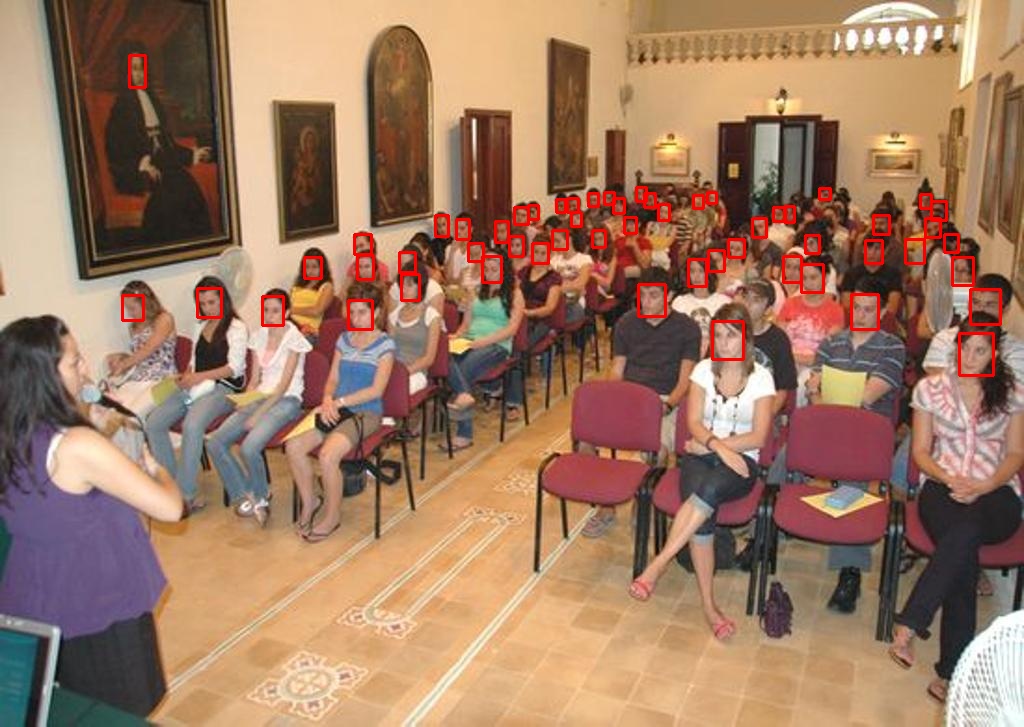} &
\includegraphics[width=0.27\linewidth]{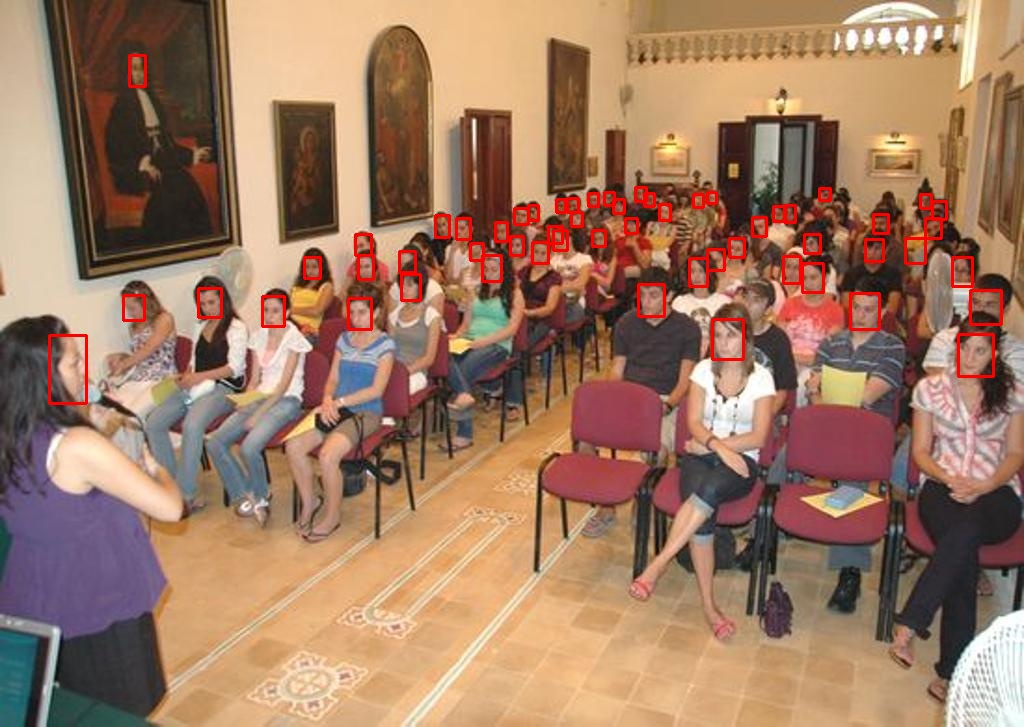} &
\includegraphics[width=0.27\linewidth]{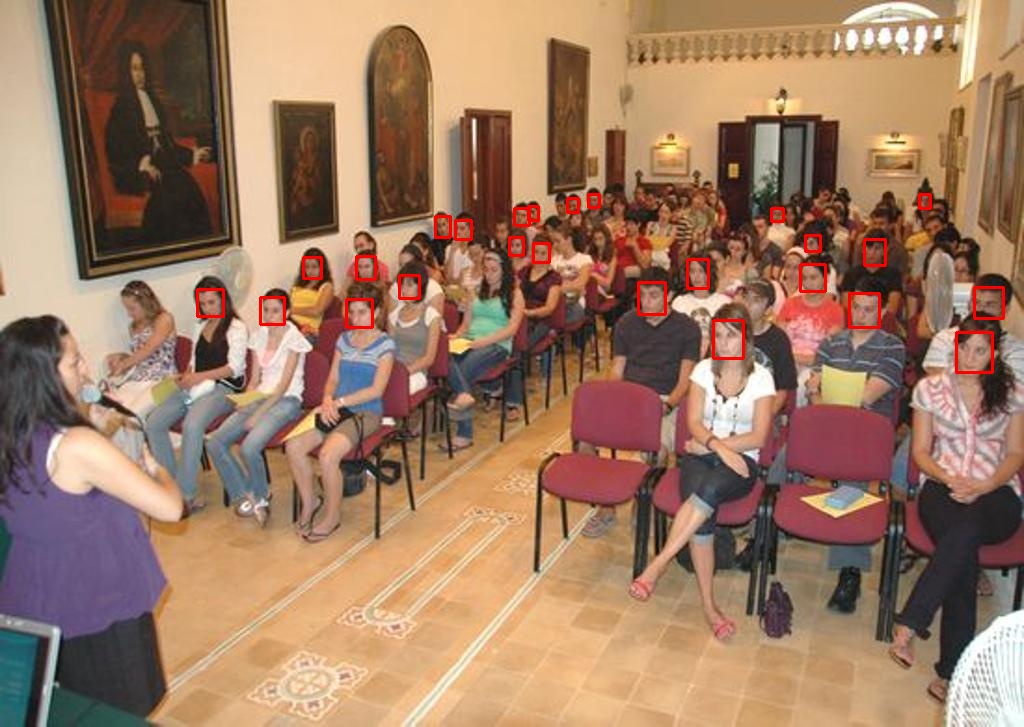}
\end{tabular}
\end{center}
\caption{Visualisation of face detection examples using the original EXTD model (first column) and its pruned variants, EXTD10 (second column) and EXTD50 (third column). We observe that for some examples, the model produced by the proposed approach with 10\% pruning rate provides improved face detection performance.   
}
\label{fig:ExamplesEXTD}
\end{figure*}

\begin{figure*}[!htb]
\begin{center}
\begin{tabular}{ccc}
EResFD & EResFD10 & EResFD50 \\
\includegraphics[width=0.27\linewidth]{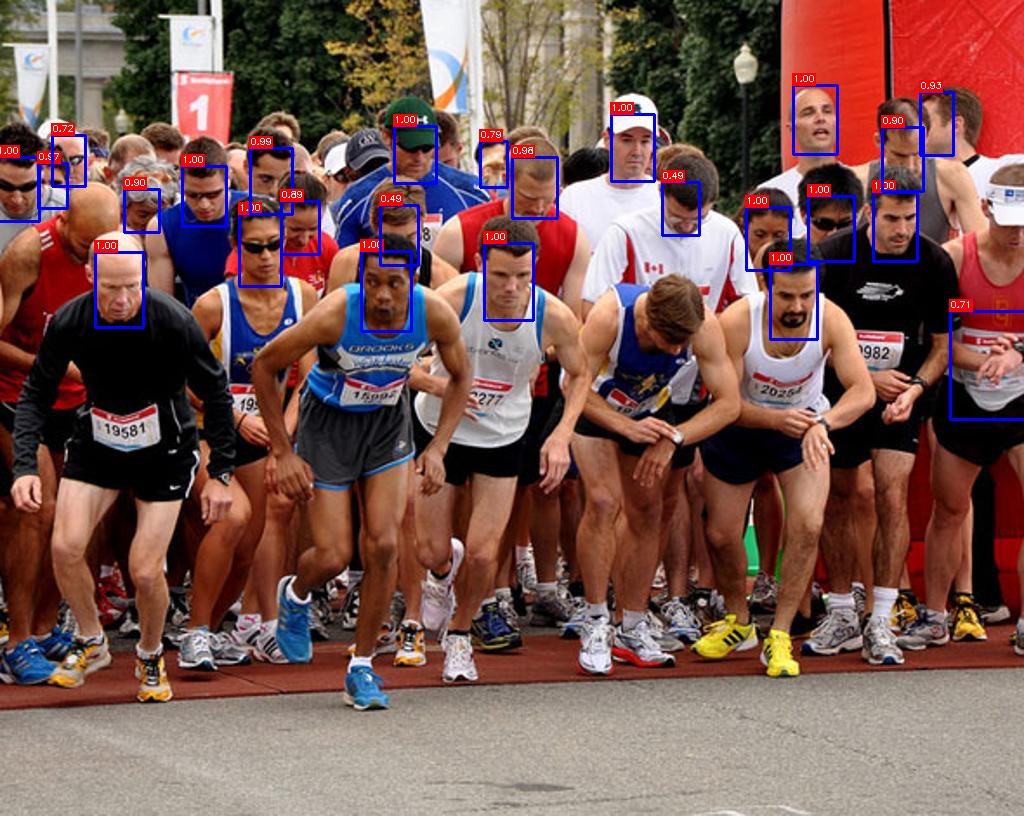} &
\includegraphics[width=0.27\linewidth]{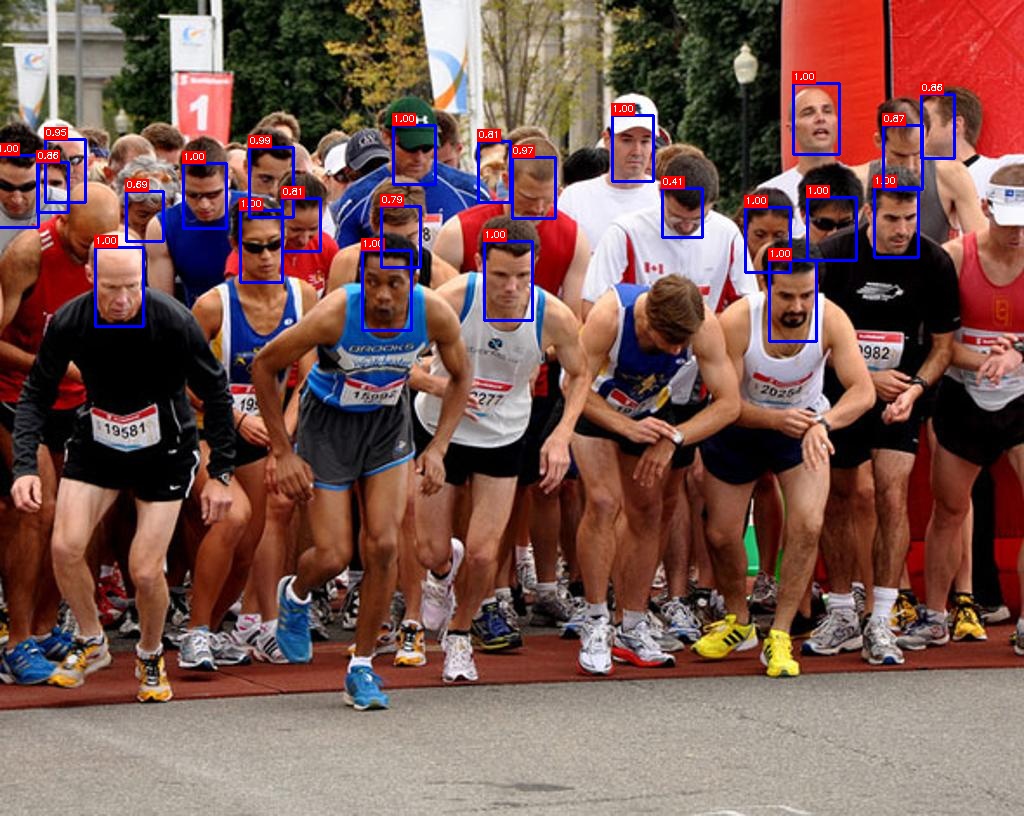} &
\includegraphics[width=0.27\linewidth]{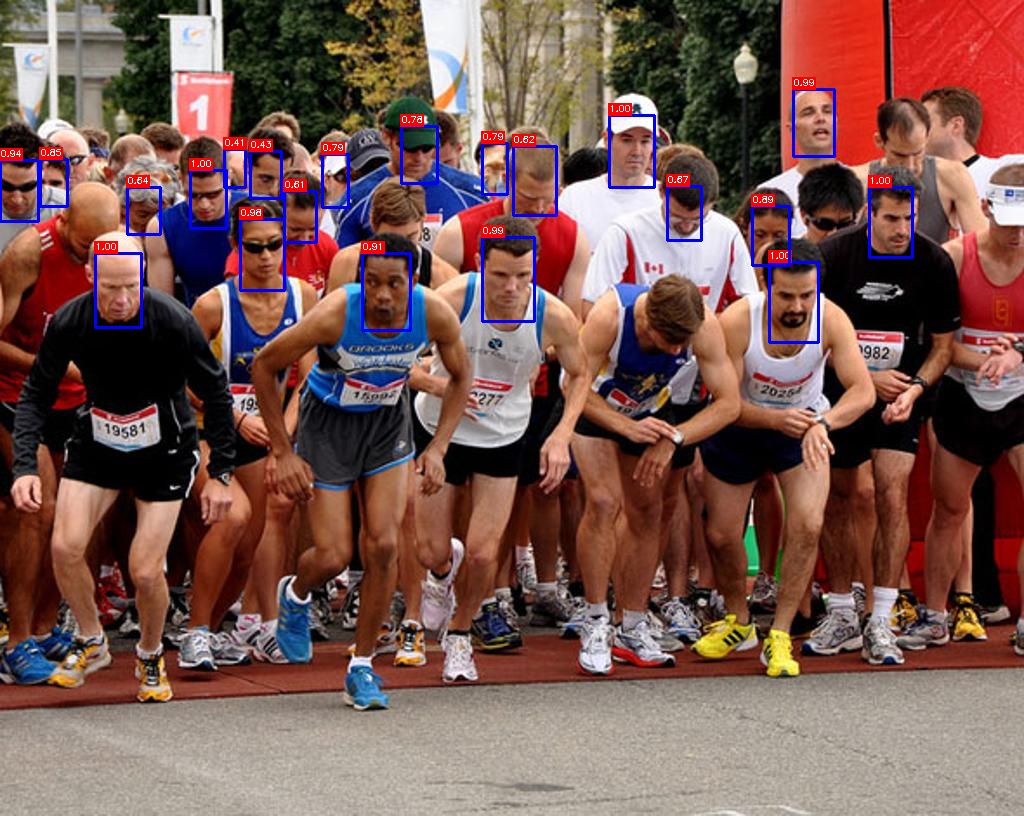} \\
\includegraphics[width=0.27\linewidth]{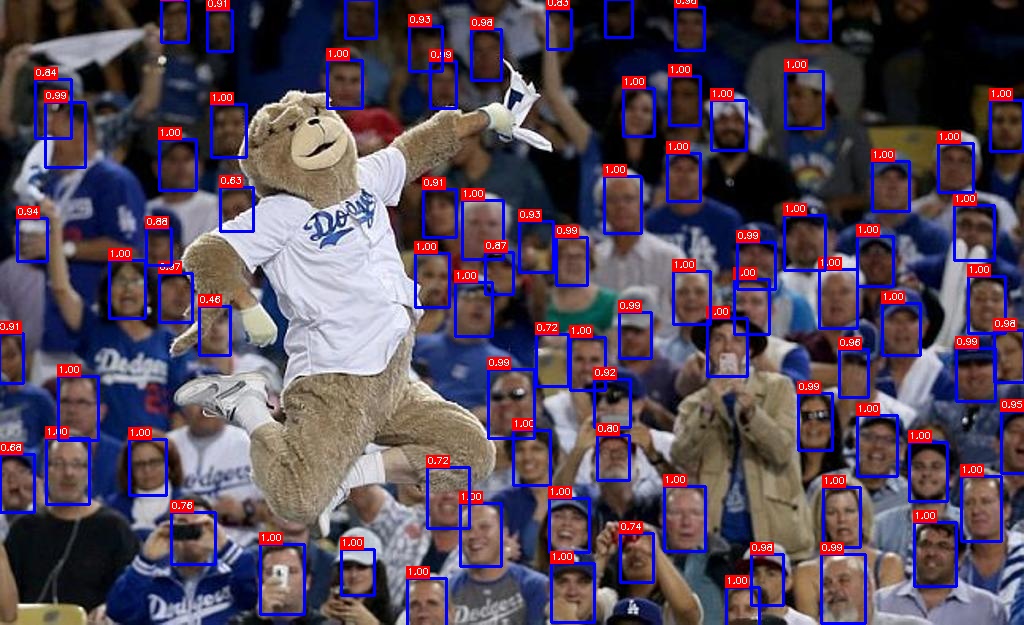} &
\includegraphics[width=0.27\linewidth]{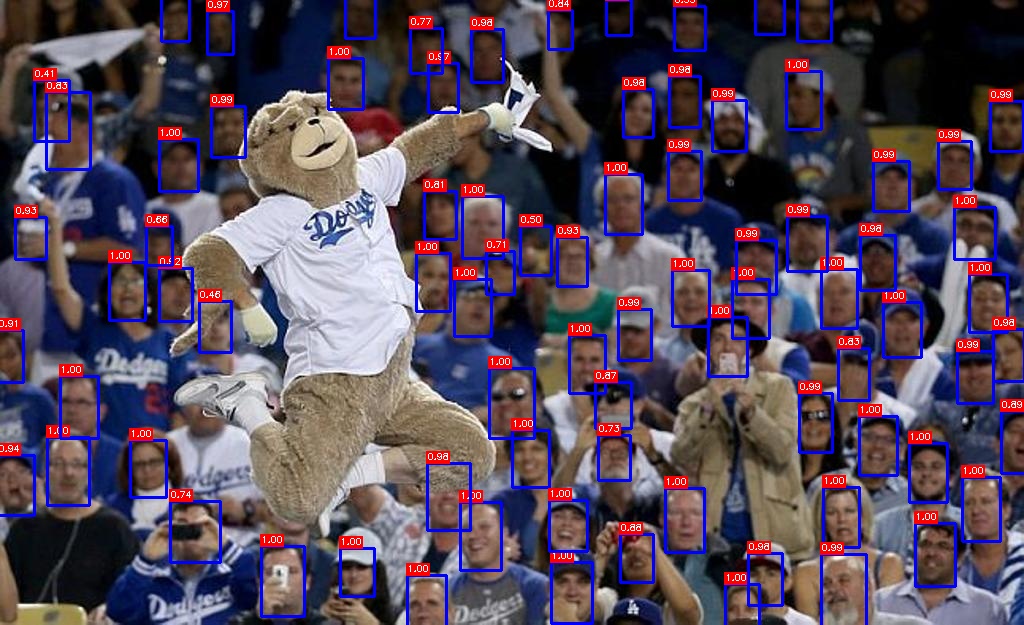} &
\includegraphics[width=0.27\linewidth]{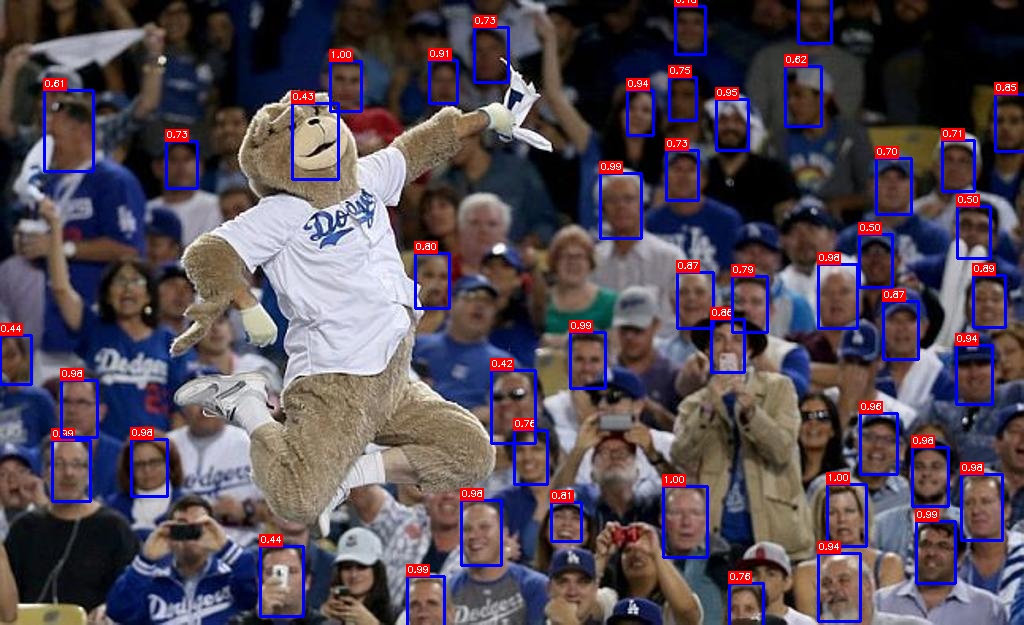} \\
\includegraphics[width=0.27\linewidth]{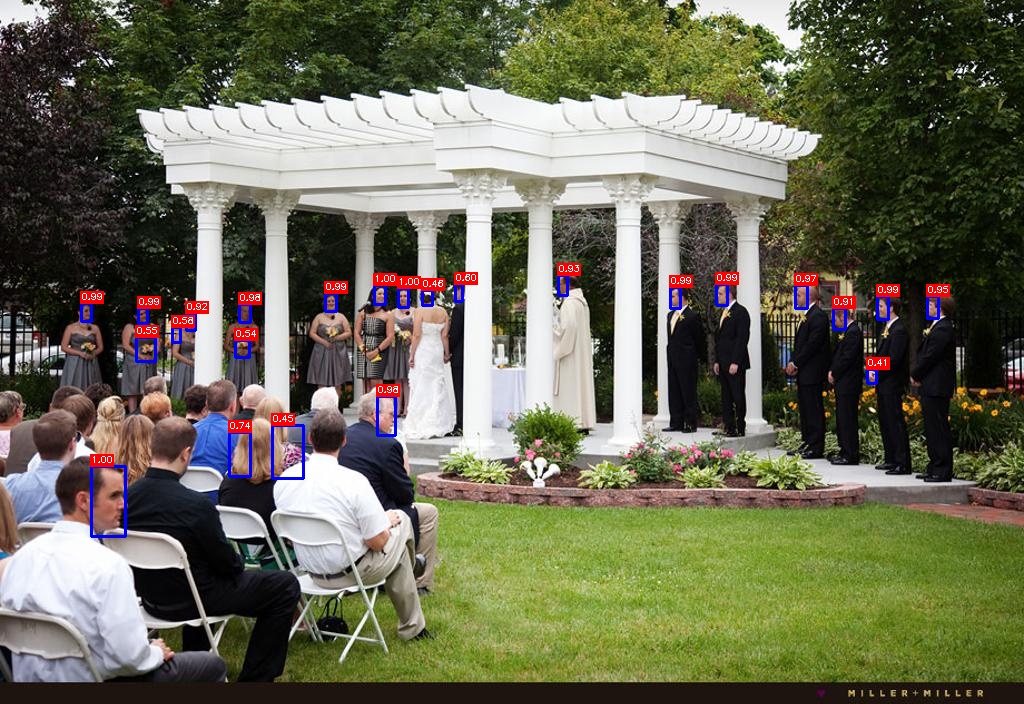} &
\includegraphics[width=0.27\linewidth]{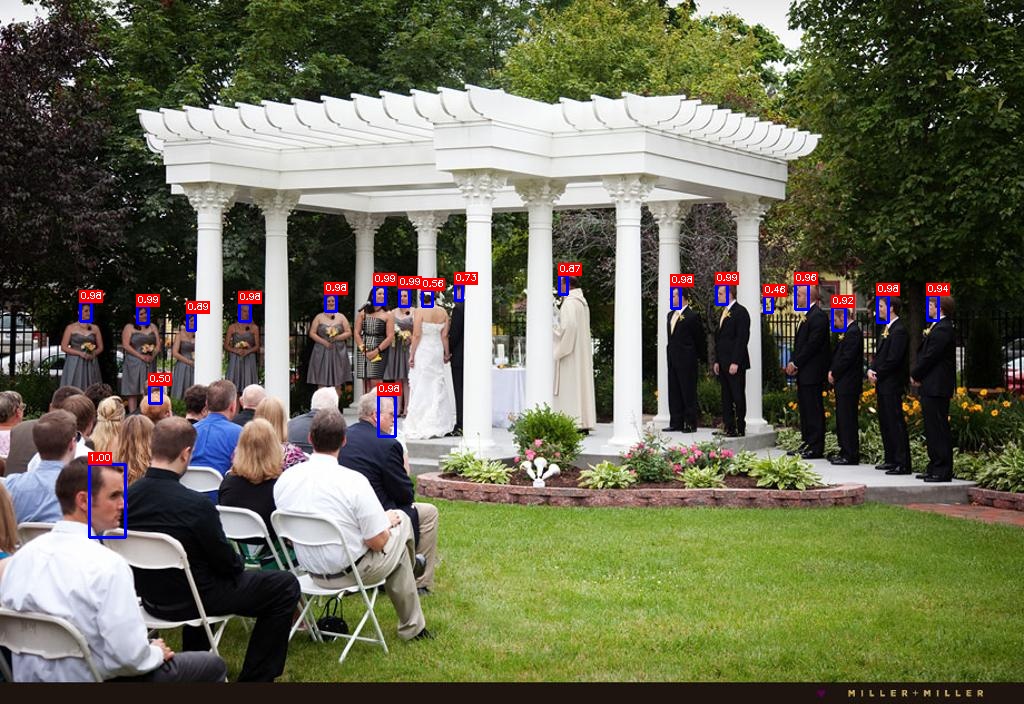} &
\includegraphics[width=0.27\linewidth]{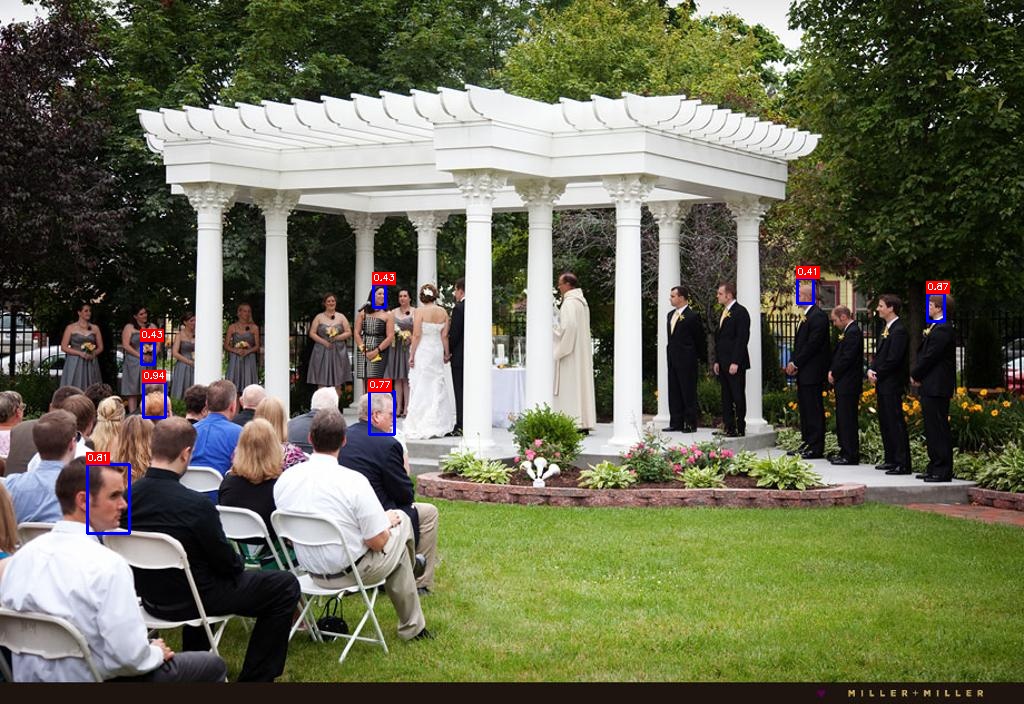} \\
\includegraphics[width=0.27\linewidth]{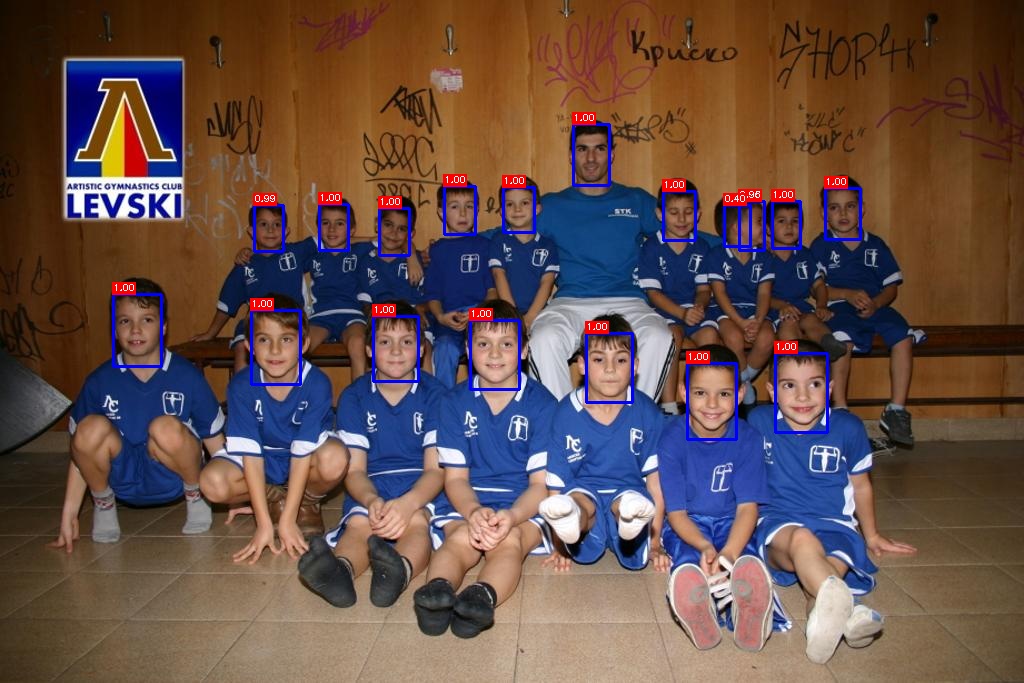} &
\includegraphics[width=0.27\linewidth]{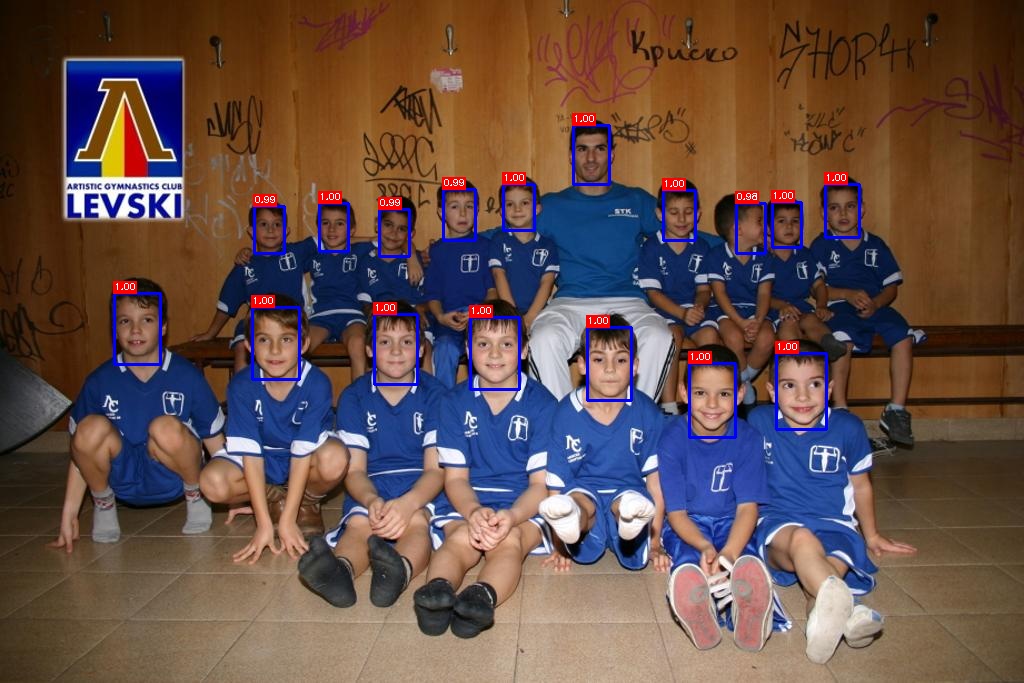} &
\includegraphics[width=0.27\linewidth]{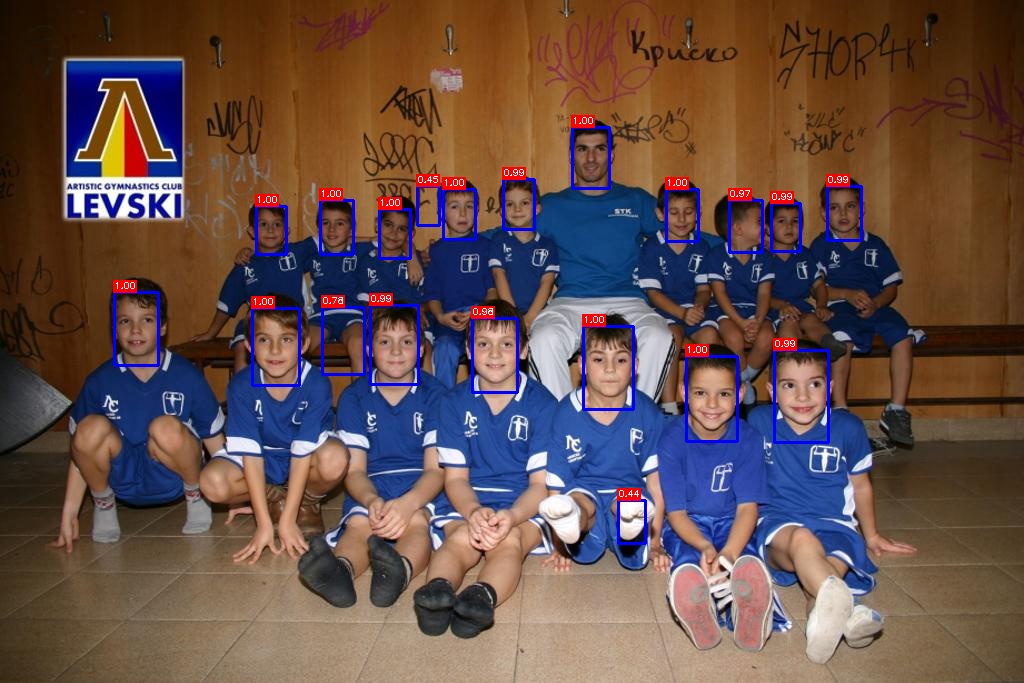}
\end{tabular}
\end{center}
\caption{Visualisation of face detection examples using the original EResFD model (first column) and its pruned variants, EResFD10 (second column) and EResFD50 (third column). As with our EXTD examples (Figure \ref{fig:ExamplesEXTD}), we see that in many cases, the model derived using the proposed approach with 10\% pruning rate outperforms the original one in terms of detection performance.
}
\label{fig:ExamplesEResFD}
\end{figure*}

\subsection{Results}
\label{ssec:results}

The pruned models produced using the proposed pruning approach are compared against the following methods that represent the state-of-the-art in face detectors (especially in lightweight ones, although the comparison is not limited to lightweight models): RetinaFace (both with MobileNet and ResNet50 as backbone) \cite{deng2019retinaface}, TinaFace \cite{zhu2020tinaface}, Yolo5Face \cite{qi2022yolo5face}, LFFD \cite{he2019lffd}, S3FD \cite{zhang2017s3fd}, Dual Shot Face Detector \cite{li2019dsfd}, MTCNN \cite{8110322}, FaceBoxes \cite{zhang2017faceboxes}, PyramidBox \cite{tang2018pyramidbox}, SCRDFD \cite{guo2021sample}, EagleEye \cite{s19092158}, PruneFaceDet \cite{10.1145/3436369.3437415}, and the original EXTD \cite{yoo2019extd} and EResFD \cite{jeong2022eresfd} models.

Figures \ref{fig:fig1} and \ref{fig:fig2} illustrate a comparative performance analysis of our pruned models against the referenced ones on the Easy and Hard WIDER FACE subsets, respectively.
We should note that in these figures, EXTD and EResFD denote our reproduced results based on training and evaluating the corresponding models, EXTD(original) and EResFD(original) refer to the scores reported in the original publications, and EXTD$\theta$ and EResFD$\theta$
refer to the models produced by pruning them with rate $\theta\%$ using the proposed approach.

Tables \ref{tab:sfp_on_extd} and \ref{tab:sfp_on_eres} compare the proposed approach against our baseline approach, i.e., the SFP procedure combined with the L1 Norm pruning algorithm, across EXTD and EResFD, and for the five different pruning rates.
In these tables, the Real Sparsity column reports the actual sparsity achieved using a specific pruning rate with the respective NNI pruner \cite{nni2021}.
More specifically, the NNI pruner optimally prunes filters within each layer to ensure that the overall model sparsity does not exceed the user-defined threshold; this leads to real sparsity values being slightly lower than the target sparsity values as defined with the pruning rate input. On the other hand, the \# of Parameters column provides the number of parameters of the model computed using a custom PyTorch \cite{paszke2019pytorch} based routine. These tables offer insights into the trade-off between model performance, sparsity and number of parameters, providing an overview of the effectiveness of the pruning strategies on those face detectors.

Finally, Figures \ref{fig:ExamplesEXTD} and \ref{fig:ExamplesEResFD} provide illustrative face detection examples of the original EXDT and EResFD models and the pruned models produced using the proposed approach with pruning rates 10\% and 50\%. From the obtained results we conclude the following:

i) The family of models produced by the proposed approach provide a competitive detection performance with significantly reduced model size.

ii) Our pruned model with 10\% sparsity exhibits a slightly improved performance when compared to our reproduced EXTD and EResFD.
This improvement can be attributed to the pruning process acting as a form of regularization.

iii) From the overall results in Tables \ref{tab:sfp_on_extd} and \ref{tab:sfp_on_eres} we observe the superiority of the proposed approach (based on FPGM algorithm) over the baseline approach  (based on the L1 Norm algorithm), across all pruning rates.

iv) A similar conclusion to the above can be drawn by observing the qualitative face detection results in the middle column of both Figures \ref{fig:ExamplesEXTD} and \ref{fig:ExamplesEResFD}.
Specifically, we see that  EXTD10 and EResFD10 are able to identify faces that would otherwise have been missed by the original model.
Furthermore, our pruned model with 10\% sparsity also manages to exclude some of the false positives that the original model produced, as it can be seen in the third row of Figure \ref{fig:ExamplesEResFD}.

v) In overall, from the presented results it is evident that as sparsity increases, there is a pronounced decline in performance within the 'Hard' subset. The drop in 'Easy' and 'Medium' subsets is more modest. By varying the pruning rate, our approach yields a family of lightweight detection models that represent different compromises between model size and detection accuracy. Furthermore, our findings suggest that FPGM has the potential to enhance the model's accuracy when enforcing small pruning rates.

\section{Conclusions}
\label{sec:conclusions}

Face detection is a rapidly evolving domain, and the demand for lightweight and efficient models that are suitable for edge devices is high. In this paper, we explored the potential of filter pruning techniques on two already compact face detectors, namely EXTD \cite{yoo2019extd} and EResFD \cite{jeong2022eresfd}. Our research focused on two pruning algorithms: L1 Norm and Filter Pruning via Geometric Median (FPGM). Through experiments, we showed the superiority of the FPGM over the L1-Norm criterion. It is worth mentioning here, that the EResFD pruned model with 10\% sparsity showcased a slight improvement in mAP, suggesting that the pruning process can act as a form of regularization. However, as sparsity increases, there is a notable decline in the mAP of the pruned models. By varying the pruning rate, we generated a family of even more compact face detectors for use in real-life applications where the model size is a critical factor.


{\small
\begin{thebibliography}{10}\itemsep=-1pt

\bibitem{Alzubaidi2021}
Laith Alzubaidi, Jinglan Zhang, Amjad~J. Humaidi, Ayad Al-Dujaili, Ye Duan, Omran Al-Shamma, J. Santamaría, Mohammed~A. Fadhel, et~al.
\newblock Review of deep learning: Concepts, {CNN} architectures, challenges, applications, future directions.
\newblock {\em Journal of Big Data}, 2021.

\bibitem{deng2019retinaface}
Jiankang Deng, Jia Guo, Yuxiang Zhou, Jinke Yu, Irene Kotsia, and Stefanos Zafeiriou.
\newblock {RetinaFace}: Single-stage dense face localisation in the wild.
\newblock {\em arXiv preprint arXiv:1905.00641}, 2019.

\bibitem{9327924}
Nikolaos Gkalelis and Vasileios Mezaris.
\newblock Structured pruning of {LSTMs} via eigenanalysis and {G}eometric {M}edian for mobile multimedia and deep learning applications.
\newblock In {\em 2020 IEEE International Symposium on Multimedia (ISM)}, pages 122--126, 2020.

\bibitem{guo2021sample}
Jia Guo, Jiankang Deng, Alexandros Lattas, and Stefanos Zafeiriou.
\newblock Sample and computation redistribution for efficient face detection.
\newblock {\em arXiv preprint arXiv:2105.04714}, 2021.

\bibitem{he2018soft}
Yang He, Guoliang Kang, Xuanyi Dong, Yanwei Fu, and Yi Yang.
\newblock Soft filter pruning for accelerating deep convolutional neural networks.
\newblock {\em arXiv preprint arXiv:1808.06866}, 2018.

\bibitem{he2019filter}
Yang He, Ping Liu, Ziwei Wang, Zhilan Hu, and Yi Yang.
\newblock Filter pruning via {G}eometric {M}edian for deep convolutional neural networks acceleration.
\newblock In {\em Proceedings of the IEEE Conference on Computer Vision and Pattern Recognition (CVPR)}, 2019.

\bibitem{he2019lffd}
Yonghao He, Dezhong Xu, Lifang Wu, Meng Jian, Shiming Xiang, and Chunhong Pan.
\newblock Lffd: A light and fast face detector for edge devices.
\newblock {\em arXiv preprint arXiv:1904.10633}, 2019.

\bibitem{howard2017mobilenets}
Andrew~G Howard, Menglong Zhu, Bo Chen, Dmitry Kalenichenko, Weijun Wang, Tobias Weyand, et~al.
\newblock {Mobilenets}: Efficient convolutional neural networks for mobile vision applications.
\newblock {\em arXiv preprint arXiv:1704.04861}, 2017.

\bibitem{jeong2022eresfd}
Joonhyun Jeong, Beomyoung Kim, Joonsang Yu, and Youngjoon Yoo.
\newblock {EResFD}: Rediscovery of the effectiveness of standard convolution for lightweight face detection.
\newblock {\em arXiv preprint arXiv:2204.01209}, 2022.

\bibitem{Kumar2019}
Ashu Kumar, Amandeep Kaur, and Munish Kumar.
\newblock Face detection techniques: a review.
\newblock {\em Artificial Intelligence Review}, 52(2):927--948, 2019.

\bibitem{Kumar2021}
Aakash Kumar, Ali~Muhammad Shaikh, Yun Li, Hazrat Bilal, and Baoqun Yin.
\newblock Pruning filters with {L1}-norm and capped {L1}-norm for {CNN} compression.
\newblock {\em Applied Intelligence}, 2021.

\bibitem{li2016pruning}
Hao Li, Asim Kadav, Igor Durdanovic, Hanan Samet, and Hans~Peter Graf.
\newblock Pruning filters for efficient convnets.
\newblock {\em arXiv preprint arXiv:1608.08710}, 2016.

\bibitem{li2019dsfd}
Jian Li, Yabiao Wang, Changan Wang, Ying Tai, Jianjun Qian, Jian Yang, Chengjie Wang, Jilin Li, and Feiyue Huang.
\newblock {DSFD}: dual shot face detector.
\newblock In {\em Proceedings of the IEEE/CVF Conference on Computer Vision and Pattern Recognition}, pages 5060--5069, 2019.

\bibitem{10.1145/3436369.3437415}
Jingsheng Lin, Xu Zhao, Nanfei Jiang, and Jinqiao Wang.
\newblock {PruneFaceDet}: Pruning lightweight face detection network by sparsity training.
\newblock In {\em Proceedings of the 2020 9th International Conference on Computing and Pattern Recognition}, ICCPR '20, page 181–186, New York, NY, USA, 2021. Association for Computing Machinery.

\bibitem{liu2016ssd}
Wei Liu, Dragomir Anguelov, Dumitru Erhan, Christian Szegedy, Scott Reed, Cheng-Yang Fu, and Alexander~C Berg.
\newblock {SSD}: Single shot multibox detector.
\newblock In {\em Computer Vision--ECCV 2016: 14th European Conference, Amsterdam, The Netherlands, October 11--14, 2016, Proceedings, Part I 14}, pages 21--37. Springer, 2016.

\bibitem{Luo_2017_ICCV}
Jian-Hao Luo, Jianxin Wu, and Weiyao Lin.
\newblock {ThiNet}: A filter level pruning method for deep neural network compression.
\newblock In {\em Proceedings of the IEEE International Conference on Computer Vision (ICCV)}, Oct 2017.

\bibitem{nni2021}
{Microsoft}.
\newblock {Neural Network Intelligence library, https://github.com/microsoft/nni}, 2021.

\bibitem{minaee2021going}
Shervin Minaee, Ping Luo, Zhe Lin, and Kevin Bowyer.
\newblock Going deeper into face detection: A survey.
\newblock {\em arXiv preprint arXiv:2103.14983}, 2021.

\bibitem{paszke2019pytorch}
Adam Paszke, Sam Gross, Francisco Massa, Adam Lerer, James Bradbury, Gregory Chanan, et~al.
\newblock Pytorch: An imperative style, high-performance deep learning library.
\newblock {\em Advances in neural information processing systems}, 32, 2019.

\bibitem{qi2022yolo5face}
Delong Qi, Weijun Tan, Qi Yao, and Jingfeng Liu.
\newblock {YOLO5Face}: Why reinventing a face detector.
\newblock In {\em European Conference on Computer Vision}, pages 228--244. Springer, 2022.

\bibitem{248452}
R. Reed.
\newblock Pruning algorithms-a survey.
\newblock {\em IEEE Transactions on Neural Networks}, 1993.

\bibitem{tang2018pyramidbox}
Xu Tang, Daniel~K Du, Zeqiang He, and Jingtuo Liu.
\newblock {Pyramidbox}: A context-assisted single shot face detector.
\newblock In {\em Proceedings of the European conference on computer vision (ECCV)}, pages 797--813, 2018.

\bibitem{8110322}
Jia Xiang and Gengming Zhu.
\newblock Joint face detection and facial expression recognition with {MTCNN}.
\newblock In {\em 2017 4th International Conference on Information Science and Control Engineering (ICISCE)}, pages 424--427, 2017.

\bibitem{yang2016wider}
Shuo Yang, Ping Luo, Chen-Change Loy, and Xiaoou Tang.
\newblock {WIDER} {FACE}: A face detection benchmark.
\newblock In {\em Proceedings of the IEEE conference on computer vision and pattern recognition}, pages 5525--5533, 2016.

\bibitem{yoo2019extd}
YoungJoon Yoo, Dongyoon Han, and Sangdoo Yun.
\newblock {EXTD}: Extremely tiny face detector via iterative filter reuse.
\newblock {\em arXiv preprint arXiv:1906.06579}, 2019.

\bibitem{zhang2017faceboxes}
Shifeng Zhang, Xiangyu Zhu, Zhen Lei, Hailin Shi, Xiaobo Wang, and Stan~Z Li.
\newblock {Faceboxes}: A {CPU} real-time face detector with high accuracy.
\newblock In {\em 2017 IEEE International Joint Conference on Biometrics (IJCB)}, pages 1--9. IEEE, 2017.

\bibitem{zhang2017s3fd}
Shifeng Zhang, Xiangyu Zhu, Zhen Lei, Hailin Shi, Xiaobo Wang, and Stan~Z Li.
\newblock {S3FD}: Single shot scale-invariant face detector.
\newblock In {\em Proceedings of the IEEE international conference on computer vision}, pages 192--201, 2017.

\bibitem{s19092158}
Xu Zhao, Xiaoqing Liang, Chaoyang Zhao, Ming Tang, and Jinqiao Wang.
\newblock Real-time multi-scale face detector on embedded devices.
\newblock {\em Sensors}, 2019.

\bibitem{sym12091426}
Shangping Zhong, Wude Weng, Kaizhi Chen, and Jianhua Lai.
\newblock Deep-learning steganalysis for removing document images on the basis of {G}eometric {M}edian pruning.
\newblock {\em Symmetry}, 2020.

\bibitem{zhu2020tinaface}
Yanjia Zhu, Hongxiang Cai, Shuhan Zhang, Chenhao Wang, and Yichao Xiong.
\newblock {Tinaface}: Strong but simple baseline for face detection.
\newblock {\em arXiv preprint arXiv:2011.13183}, 2020.

\end{thebibliography}
}

\end{document}